# SEAM: Submap-Anchored Evidence for Lifelong LiDAR Mapping under Trajectory Deformation

Kyuwon Kim[1]

*Abstract*— **We propose SEAM, a LiDAR-based lifelong mapping framework. Instead of relying on a single anchor spanning the entire session, SEAM generates evidence based on a trajectory optimized with submap-level anchors, and performs dynamic object removal and change detection. Through submap-level reprojection, the generated evidence remains usable even if the trajectory is subsequently modified by a new session, eliminating the need to recompute the entire process from scratch. SEAM suppresses geometrically unreliable inter-session loop edges using a DOP-based confidence measure. Suppressing unreliable loop edges prevents alignment errors. SEAM also uses a directional voxel-wise evidence model. The model accounts for occupancy patterns that vary with ray direction. Direction-aware evidence separates dynamic objects from environmental changes more precisely. Experiments on a real construction-site dataset and a long-term multi-session dataset show that SEAM achieves higher accuracy and faster processing than existing methods.**

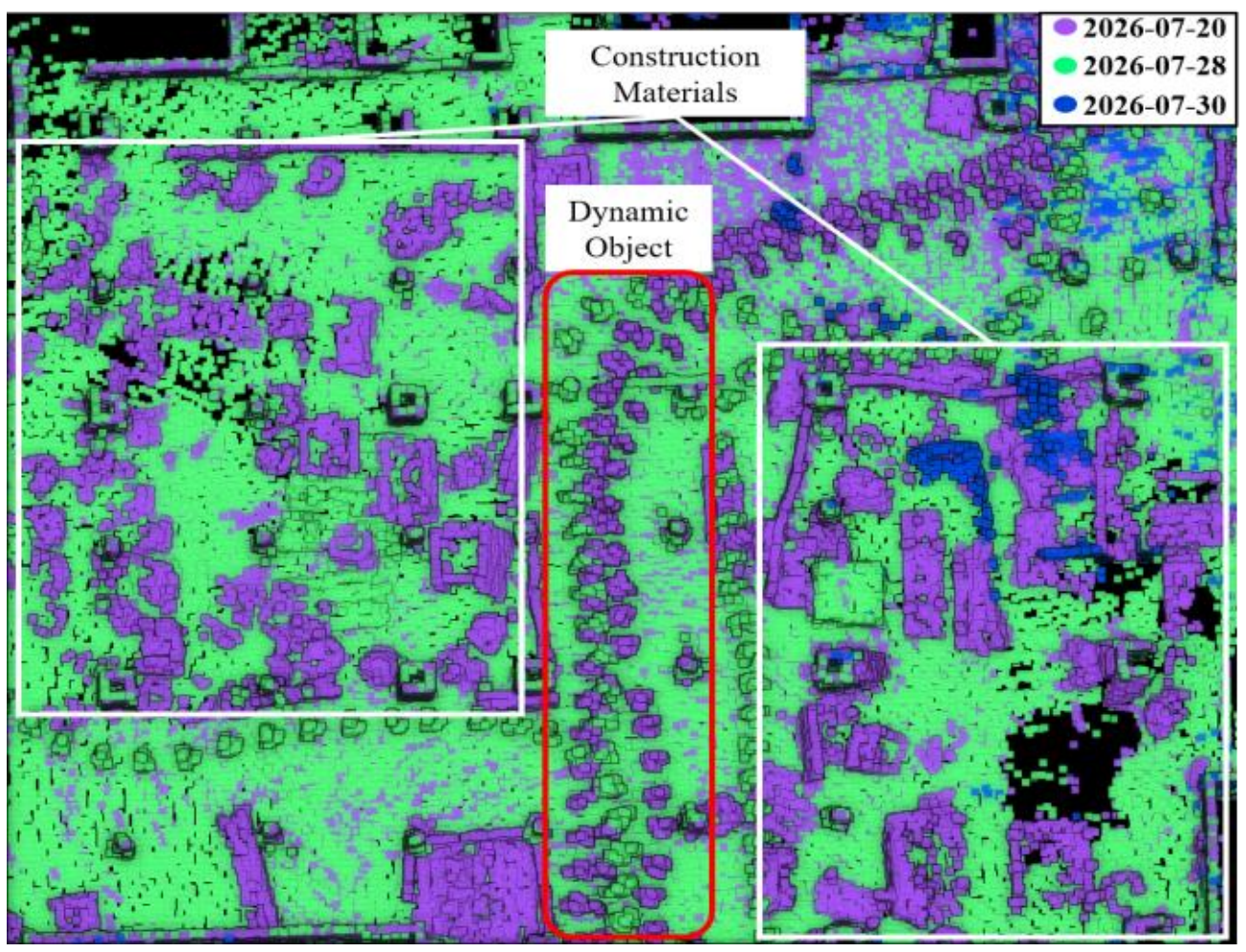

Fig. 1. An example of a construction site. Three sessions collected over ten days are overlaid, each shown in a different color. The white boxes mark construction materials that appear or disappear between visits. The red box marks dynamic objects, such as workers and equipment, that were present only momentarily yet remain as ghost points.

## I. Introduction

Light Detection and Ranging (LiDAR)-based mapping [1–5] has become increasingly important with the advancement of Autonomous Mobile Robot (AMR) technology. Demand for AMR technology continues to grow. Among LiDAR-based mapping approaches, lifelong mapping [6-8] detects environmental changes over time and updates the map accordingly. Lifelong mapping has been widely studied to support long-term robot operation. Lifelong mapping is particularly essential in environments such as construction sites. In construction sites, materials and equipment frequently change location, and new structures are continuously built. Fig. 1 shows an overlay of maps collected over 10 days at an actual construction site. As shown in the figure, dynamic objects such as workers and equipment frequently move around the site. Construction materials also appear and disappear over the 10-day period. Autonomous navigation in such environments requires the robot to merge newly acquired sessions into the existing map and detect environmental changes.

LiDAR-based lifelong mapping consists of three components: multi-session alignment [9, 10], dynamic object removal (DOR) [11–13], and change detection (CD) [14, 15]. Multi-session alignment aligns sessions acquired in different coordinate frames into a common frame. DOR filters out ghost points left by dynamic objects within a single session. CD detects changes across sessions and updates the map into a static map. Prior work has assumed that the trajectories of previously registered sessions remain fixed during map merging and updating. The fixed-trajectory assumption may hold for two or three sessions. In environments that require frequent map updates, such as construction sites, residual errors from session alignment are not corrected. Uncorrected residual errors continue to accumulate and lead to map inconsistency. Therefore, whenever a new session arrives, the trajectories of all sessions must be optimized jointly [16-18]. Joint optimization corrects the errors of previous sessions as well as the new session. However, optimizing the full set of trajectories also changes the trajectories of previous sessions. As a result, previously computed change detection results are invalidated and must be recomputed from scratch.

To address this problem, we propose a lifelong LiDAR mapping framework which reuses evidence from previous sessions at the submap level. Conventional session-anchor-based optimization fails to effectively correct drift errors of varying magnitude, because only a single anchor exists for the entire session trajectory. In contrast, submap-level anchoring partitions the trajectory into locally rigid segments (submaps) based on cumulative travel distance and assigns an independent anchor variable to each segment. As a result, drift of varying magnitude and direction can be individually corrected on a per-segment basis. Through ray casting, we accumulate per-voxel hit and free direction statistics and the number of observing keyframes as evidence. We incorporate the resulting confidence weights into the local-global ephemerality structure of ELite [19]. Instead of point-wise ephemerality propagation, we perform DOR and CD using a confidence weight computed from the per-voxel ray-direction statistics and the number of observing keyframes. The evidence is stored at the submap-anchor level and is reprojected via an SE(3) transformation to match the re-optimized trajectories. Submap-anchor based reprojection allows the evidence to be reused without recomputation for every session, reducing map update time.

[1]Kyuwon Kim is with the HMG Construction R&D Division, Hyundai Engineering & Construction, Seoul, Republic of Korea, 03058 (e-mail: kkw1125@hdec.co.kr).

The main contributions of this paper are as follows:

- We introduce SEAM, a lifelong LiDAR mapping framework. SEAM anchors the pose graph at the submap level to support joint re-optimization of all sessions and uses the resulting anchors to reproject stored evidence for DOR and CD.
- We propose a LiDAR FoV-normalized DOP-based confidence criterion for inter-session loop edges which prevents erroneous edges and the resulting error propagation across the session trajectory.
- We propose a direction-aware voxel evidence model for DOR and CD. The model distinguishes reliable multi-directional observations and reuses prior evidence through submap-anchor-based reprojection to reduce map update time.
- We demonstrate SEAM on a real construction-site dataset and a long-term dataset. SEAM jointly merges more than three sessions while faithfully capturing environmental changes.

## II. Related Works

### A. Multi-Session Map Alignment

In multi-session alignment, each session has a different reference coordinate system. Aligning sessions requires a process that maps them into a single common coordinate system [20]. Previous studies have detected inter-session loops via keyframe-level place recognition [21-23] or matching between session maps [24]. Such work used the detected loops as constraints within anchor node-based pose graph optimization [25]. However, most such methods designate an already aligned session as the central session. Such methods then sequentially accumulate newly arriving sessions one at a time. Joint re-optimization of three or more sessions at once is rare among such methods.

### B. Dynamic Object Removal

Research on removing dynamic objects within a single session has represented the environment using various methods, such as voxels, range images, and bins. OctoMap [11] determines voxel occupancy probability via ray tracing. DUFOMap [12] identifies void regions through single-scan ray casting. Removert [26] and ERASER [13] evaluate geometric discrepancies against an existing map with multi-resolution range images and a polar-coordinate pseudo-occupancy descriptor. In contrast, learning-based methods [27, 28] directly predict the dynamic status of individual points. However, these methods require large-scale training data. Conventional occupancy-based methods accumulate hit and free observations as scalar probabilities. Scalar-probability accumulation treats surfaces observed from multiple directions the same as surfaces observed repeatedly from a single direction. Grazing incidence causes reflection instability. As a result, conventional methods misread the spurious free observations as non-existent surfaces and deletes valid structure.

### C. Change Detection

Change detection between sessions falls into two categories. One category of methods assumes perfect map alignment and extracts only differences [29]. The other category assumes that alignment errors may exist and directly compares the new session with the existing map [30]. LT-mapper [31] distinguishes between high- and low-dynamic objects using range images of varying resolutions. It removes and then reconstructs points that have changed, but represents the results solely as binary values indicating appearance or disappearance. ELite [19] introduced a two-stage probabilistic ephemerality model consisting of local ephemerality and global ephemerality. The two-stage model finely represents the gradual transition between the two states. Voxel-based methods generally suffer from static and dynamic points being mixed within a single voxel. ELite avoids the mixing problem by propagating ephemerality through point-based nearest-neighbor propagation. However, ELite is required very long processing times because of point-based propagation. MTD-Map [15] integrates DOR and CD into a single stage by modeling the direction and duration of occupancy transitions. MTD-Map uses a voxel-wise bipolar representation to distinguish between arriving and vacating events without point-based propagation. The bipolar representation greatly reduces processing time. However, MTD-Map has the limitation of presupposing already matched trajectories without a separate inter-session matching process. All the above methods assume that the trajectories of previous sessions are fixed. The fixed-trajectory assumption causes such methods to fail to account for trajectory deformations that occur when multiple sessions are re-optimized simultaneously.

## III. SEAM: Submap Anchored Map Alignment and Directional Evidence based Map Update

### A. System Overview

As shown in Fig. 2, the SEAM pipeline covers the entire lifelong mapping process. The pipeline includes multi-session map alignment, dynamic object removal, and change detection. When a new session begins, the new session's trajectory is segmented into submaps based on travel distance. Multi-session map alignment aligns the coordinate systems of the new session and the previous sessions. Global registration is performed first between the map accumulated from prior sessions and the new session's map. Inter-session loop detection then identifies loops across nodes from all sessions and establishes edges. Finally, the trajectories of all sessions are jointly optimized. The new session generates directional evidence from the optimized trajectory and scan data. SEAM removes dynamic objects by computing a local ephemerality value $\varepsilon_l$ from the directional evidence. For the previous sessions, SEAM reprojects the evidence and the previously generated dynamic-object-removed maps via an SE(3) transformation, using the re-optimized submap anchors. Finally, the refined map of each session is compared against the map accumulated from prior sessions. The comparison classifies points into five categories, which update the global ephemerality value $\varepsilon_g$ . The updated global ephemerality value produces a lifelong map that retains $\varepsilon_g$ . The same value also produces a static map that keeps only the static structures filtered by $\varepsilon_g$ . Submap anchors and evidence are reused whenever the map of a subsequent session arrives. As sessions accumulate, reusing the costly evidence-generation and DOR steps significantly reduces processing time.

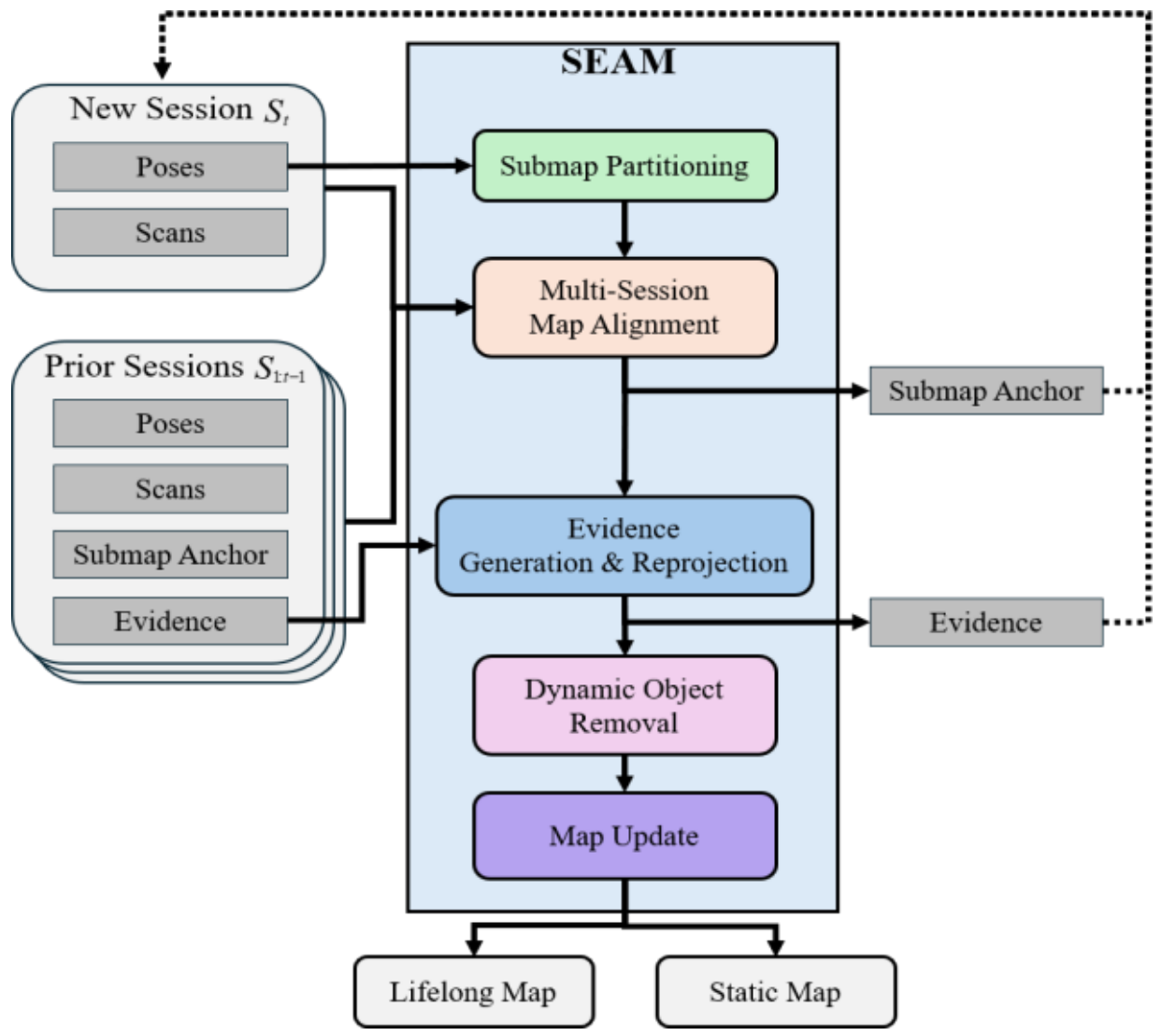

Fig. 2. Overview of the SEAM system pipeline.

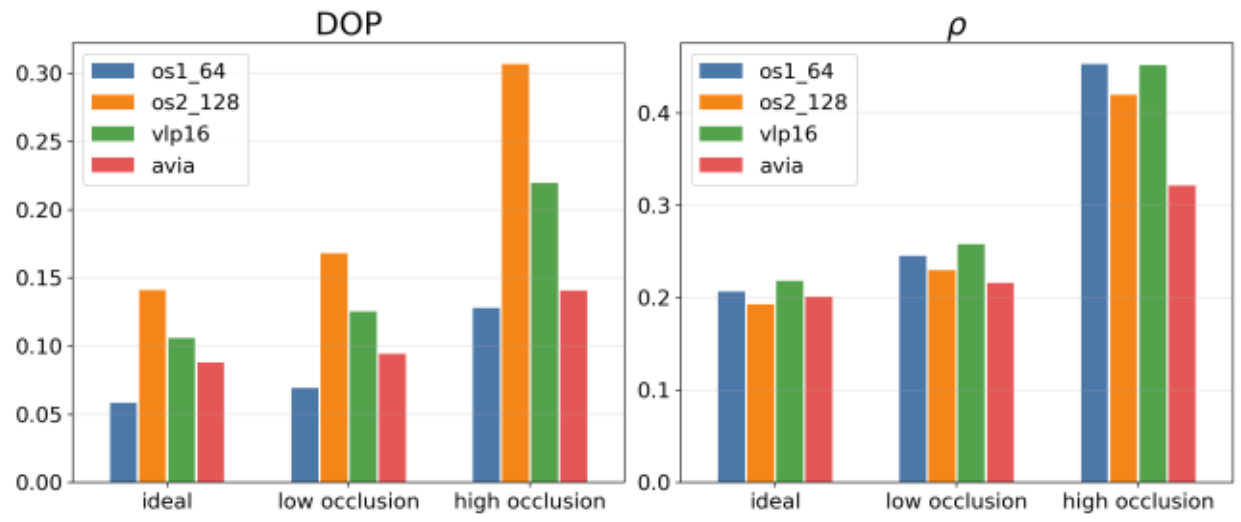

Fig. 3. Comparison of DOP and $\rho$ across different LiDAR models under varying occlusion levels.

### B. Submap Anchor based Map Alignment

When a new session arrives, alignment with the previous sessions starts. First, a map-to-map coarse alignment [10] is performed between the new session's map and the static map accumulated from the previous sessions. LT-Mapper [31] performs optimization directly on the edges built from place recognition based inter-session loop detection, without a coarse alignment step. However, if the loops are insufficient or the discrepancy between the two coordinate frames is large, convergence of the optimization can become unstable. Our pipeline therefore performs the coarse alignment first to obtain a stable initial estimate. The resulting relative pose becomes the initial value of the new session's submap anchor.

After the coarse alignment, SEAM performs inter-session loop detection. For each node of the new session, SEAM finds the nearest node among the previous session's nodes. The two nodes are scan-matched using Generalized Iterative Closest Point (GICP) [32] to estimate their relative pose. The reliability of the resulting match is evaluated to determine whether the edge is accepted [33]. Matching reliability is judged using a value that quantifies the geometric distribution of the 3D LiDAR point cloud. The value is inspired by Dilution of Precision (DOP) [34] from the GNSS field. Existing methods [35] only reflect the residual between correspondence points. DOP quantifies whether the matched point clouds are evenly distributed across diverse directions or whether the distribution is skewed due to occlusion. However, the Field of View (FoV) differs by LiDAR model. Therfore, normalization for the FoV difference is required.

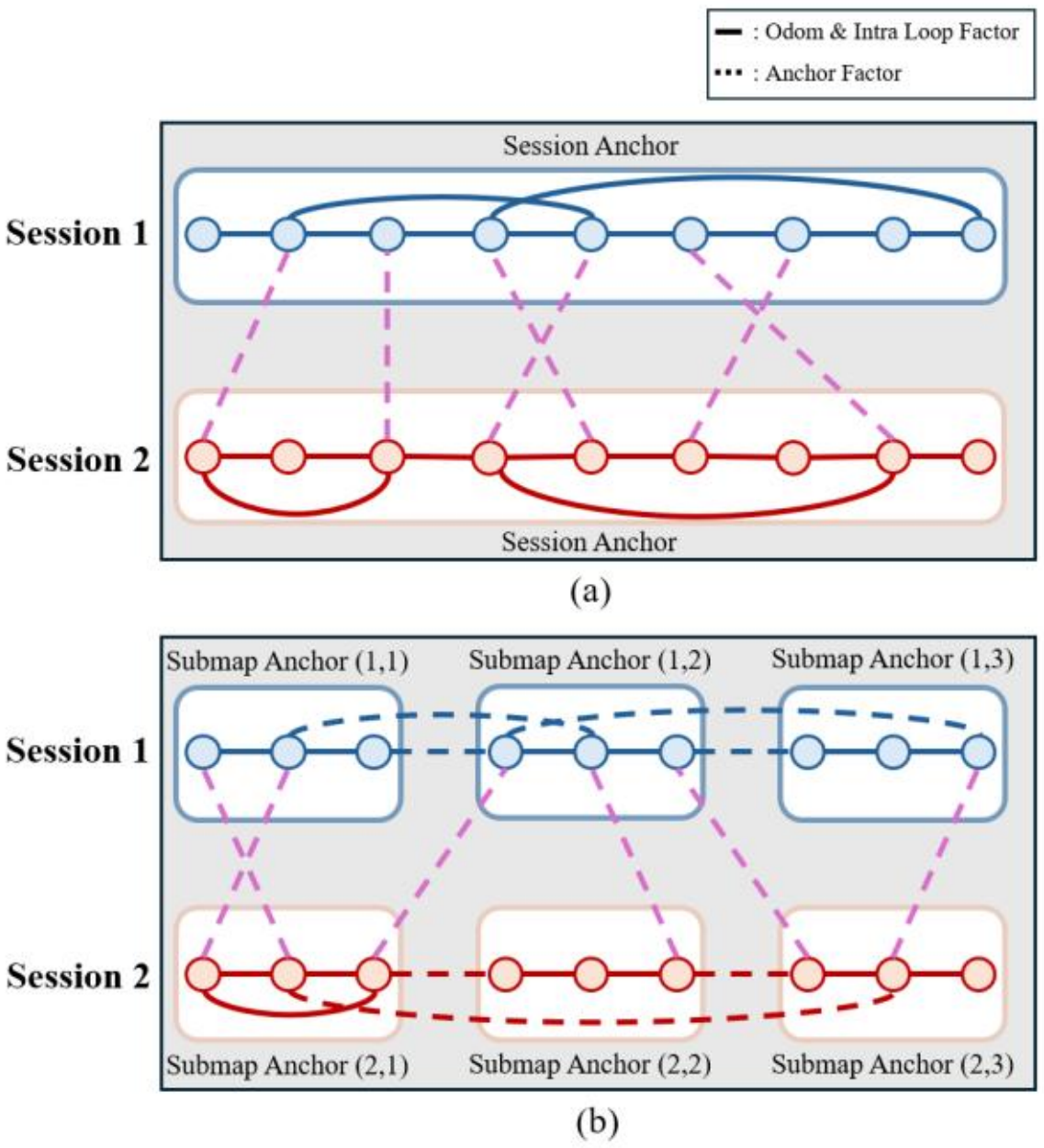

Fig. 4. Comparison of factor graphs structures of the session and submap anchor across two sessions. Solid lines denote odometry and intra-submap loop factors and dashed lines denote anchor factors. (a) Session anchor: a single anchor connects each session as a whole. (b) Submap anchor: multiple anchors connect corresponding submaps.

$$\rho = \frac{dop \cdot \sqrt{N_{typ}}}{g_0} . \quad (1)$$

$\rho$ is a normalized value of DOP. DOP can be expressed as $\sqrt{tr(G^{-1})}\Big/\sqrt{N}$ [36], where $G$ is the matrix representing the distribution of direction vectors between the sensor origin and the points and $N$ is the number of points. DOP is computed after downsampling the point cloud into voxels. Therefore, $N$ depends on the LiDAR's FoV. To remove the effect of FoV, DOP is normalized using $g_0$ and $N_{typ}$ . $g_0$ is the value of $\sqrt{tr(G^{-1})}$ under the assumption of a uniform point distribution. $N_{typ}$ is computed by taking the spherical surface area, within the solid angle covered by the FoV, at the LiDAR's mean scan range, and dividing it by the voxel size used for downsampling in the DOP computation.

Fig. 3 compares DOP and $\rho$ for different LiDAR models under various occlusion conditions. DOP shows a large numerical difference between LiDAR models even under the same conditions. The DOP gap becomes more pronounced under high occlusion. In contrast, $\rho$ removes the effect of FoV. The normalized values become nearly consistent across models. Matching reliability is judged from the ratio between the $\rho$ computed for the scan point cloud and the corresponding points after matching. Only results below the threshold are reflected as an edge.

Once all factors connecting different sessions have been detected through inter-session loop detection, global optimization is performed on the entire factor graph. Fig. 4 compares the factor graph structures of the session anchor approach and the submap anchor approach. Solid lines represent the odometry and intra-loop factors between consecutive keyframes. Dashed lines represent anchor factors. In (a), only the inter-session loop factors that connect different sessions are included. In (b), the odometry factors of consecutive submaps and intra-loop factors are additionally included. The session-anchor scheme corrects the entire session with only a single anchor. A single anchor cannot sufficiently capture trajectory deformation occurring across a large-scale session. To address this limitation, the anchor is subdivided at the submap level. Submap-level anchoring corrects even local trajectory deformation individually.

$$h_{ij}(p_i, p_j, \Delta_{m(i)}, \Delta_{m(j)}) . \quad (2)$$

$h_{ij}$ is the function obtained by transforming the keyframe poses $p_i$ and $p_j$ from the submap coordinate frame to the global coordinate frame using the anchor nodes $\Delta_{m(i)}$ and $\Delta_{m(j)}$ of their corresponding submaps. $h_{ij}$ computes the relative transformation between the two resulting global poses. The residual between the actual measurement $z_{ij}$ and $h_{ij}$ is defined as $e_{ij}$. The cost function of the anchoring factor is given as the Gaussian likelihood of the residual.

$$\phi(p_i, p_j, \Delta_{m(i)}, \Delta_{m(j)}) \propto exp(-\frac{1}{2} e_{ij}^T \Sigma_{ij}^{-1} e_{ij}) . \quad (3)$$

If the two poses belong to the same submap, the anchor-node terms cancel out. The factor then becomes identical to an ordinary between factor. If they belong to different submaps, $h_{ij}$ becomes a function of all four variables. The poses and the anchor nodes are then jointly corrected during optimization. During optimization, a strong prior factor is assigned to the anchor node of the first submap of the initial session to fix the origin of the global coordinate frame. Every submap anchor node added afterward is globally aligned relative to the global coordinate frame. Each submap's trajectory is adjusted incrementally.

### *C. Directional Evidence based Map Update*

Once all session trajectories are aligned, evidence generation and reprojection are performed to update the map. A conventional occupancy grid is represented as a scalar probability value in log-odds form. Log-odds accumulation loses information about how many times an observation was made and from which directions the observation came. Losing direction and count information makes it impossible to distinguish a handful of observations that merely grazed a wall from observations that were repeatedly verified from multiple directions. As a result, dynamic and static points become mixed within a single voxel and conventional occupancy grids cannot distinguish between them. To address this problem, we store a matrix that accumulates the direction vectors of the rays passing through each voxel, together with the number of observations made from different keyframes. We store the direction vectors and observation counts independently as hit evidence and free evidence. The accumulated hit and free evidence are used to assess the reliability of each evidence type.

$$q_i = exp(-max(0, r_i - r_0) / r_s) . \quad (4)$$

To generate evidence, weights for the rays passing through each voxel are calculated using ray casting. $q_i$ is a weight calculated based on the distance between the sensor and the ray passing through the voxel. $r_i$ denotes the distance between the voxel and the sensor. Attenuation begins once the distance exceeds $r_0$ and $r_s$ determines the scale of the attenuation. To determine whether the ray has been observed from multiple directions, the directional dispersion matrix $M$ is constructed by accumulating $q_i u_i u_i^T$. Next, using the eigenvalues $\lambda_2 \geq \lambda_1 \geq \lambda_0$ obtained by eigen-decomposing $M$. The anisotropy $\alpha$ is determined via $\lambda_2 / \lambda_1$ [37]. The anisotropy value indicates how uniformly the voxel's rays have been observed in terms of direction.

$$\omega = clip(\frac{\alpha}{\alpha_0}, 0, 1) \cdot min(1, \frac{Q}{n_0}) . \quad (5)$$

$\omega$ is the directional confidence calculated for the hit and free evidence within a voxel, respectively. Here, $\alpha/\alpha_0$ indicates whether the observations are directionally uniform, while $Q/n_0$ indicates whether the distance attenuated effective observation count is sufficient. Once each ratio exceeds its corresponding threshold, it is saturated so that reliability no longer increases further. Multiplying and combining $\alpha/\alpha_0$ and $Q/n_0$ prevents free and hit determinations from being distorted by an isotropic shape accidentally produced by a small number of observations. The combined term also prevents distortion from a large volume of biased evidence produced by grazing rays.

$$h = n_{hit} \cdot \omega_{hit}, \quad f = n_{free} \cdot \omega_{free} . \quad (6)$$

$$s = min(n_{hit} + n_{free}, n_{sat}) / n_{sat} . \quad (7)$$

$$\varepsilon_l = s \cdot \frac{f}{h + f} . \quad (8)$$

Based on the evidence generated in this way, the intra-session local ephemerality $\varepsilon_l$ is derived. The observation counts at each keyframe, $n_{hit}$ and $n_{free}$ are each multiplied by the directional reliability $\omega$ to produce the evidence amount for hit and free, respectively. In addition, we derived $\varepsilon_l$ further multiplied by the sample sufficiency $s$ and the free ratio. Through this, we prevent evidence that is directionally biased or has insufficient samples from having an excessive influence on the dynamic object determination. As a result, $\varepsilon_l$ is estimated based on reliable evidence.

The evidence generated as above is produced in alignment with the trajectory that was optimal at that point in time. If additional sessions are later added, the entire multi-session trajectory is re-optimized and deformed. Trajectory deformation causes a mismatch between the position of the stored evidence and $\varepsilon_l$ . Therefore, reusing the previous evidence requires a coordinate transformation based on the submap anchor.

$$T_i^{rigid} = T_{now,i} \cdot T_{prev,i}^{-1} . \quad (9)$$

$$\Delta T_i = A_{now,m} \cdot T_i^{rigid} \cdot A_{prev,m}^{-1} . \quad (10)$$

$$\hat{p} = \Delta T_i \cdot p . \quad (11)$$

An error arises if the reprojection transformation matrix $\Delta T_i$ is constructed using only the difference between the optimized submap anchor and the previous submap anchor. Because the internal local pose varies depending on the optimization result. $T_i^{rigid}$ is the residual correction transformation that rigidly aligns the current run's internal local pose $T_{now,i}$ within the submap to the local pose of the previous run $T_{prev,i}$ . Inserting $T_i^{rigid}$ between the anchor differences moves the previously stored evidence point $p$ to its accurate position $\hat{p}$ . In addition, $T_i^{rigid}$ reduces the seams at adjacent submap boundaries.

Once evidence generation and reprojection are complete, voxels are classified into five categories for computing $\varepsilon_g$ , following Elite [19]. ELite classifies voxels based on the nearest-neighbor distance between point sets for the two maps. The proposed method classifies voxels based on the voxel's observations and whether the voxel has an exploration history from a previous session. In all cases, $\varepsilon_g$ is updated in the same manner as Elite, except that the point-density term $\gamma$ is replaced with the directional reliability $\omega$ from (5).

## IV. EXPERIMENT

To validate the performance of the proposed framework, we used two datasets. The first dataset combines MulRan [38] and HeLiPR [39], which involves large-scale driving with long time intervals. The second dataset was acquired from real-world construction sites, where dynamic objects are abundant and environmental changes occur frequently. All experiments were conducted on a laptop, equipped with an Intel Core i7-13260H (10 cores, 16 threads, 4.9GHz) and 32GB RAM.

TABLE I. RESULT OF MATCHING RELIABLITY EVALUATION SUCCESS RATE BETWEEN VARIOUS LIDAR SENSORS

| Group | Success Rate (%) | | |
|---|---|---|---|
| | Far | Near | Near-Mismatch |
| Ouster-Ouster | 100% | 100% | 100% |
| Velodyne -Velodyne | 98% | 100% | 100% |
| Avia-Avia | 98% | 100% | 100% |
| Ouster-Avia | 100% | 94% | 100% |
| Ouster-Velodyne | 100% | 100% | 100% |

### A. Multi-Session Map Alignment

To evaluate the matching reliability determination performance across inter-session loops, we used LiDAR data from the HeLiPR dataset. The HeLiPR dataset includes sequences acquired by multiple LiDAR sensors with different FoV characteristics: Velodyne, Ouster, and Livox Avia. Sequences from three different sensor types allow us to jointly verify matching reliability determination performance across heterogeneous sensors. The evaluation was based on the distance between the ground-truth positions of each scan pair. We verified performance using 100 overlapping cases of near range scan data within 2 m, covering both matching success and matching failure cases. We also verified performance using 100 overlapping cases of far range scan data beyond 200 m. The LiDAR FoVs are $360° \times 22.5°$ for Ouster, $360° \times 15°$ for Velodyne, $70.4° \times 77.2°$ for Avia.

The results of Table I show that the proposed matching reliability determination method operates with generally high reliability even across heterogeneous LiDAR sensor combinations. Both Ouster and Velodyne are mechanical LiDARs that scan the full surrounding area. Full-surrounding scanning secures a sufficient overlapping region between the two scan datasets, resulting in a high success rate. On the other hand, Avia is a solid-state LiDAR. The horizontal FoV of Avia is narrower than that of the two mechanical LiDARs, while the vertical FoV of Avia shows a relatively wide scan pattern. The FoV-characteristic difference makes the overlapping region relatively small. A small overlapping region is expected to cause some of the observed failures.

Fig. 5 compares the map alignment results across five sessions. In (a), the trajectory optimized in the previous session is fixed. In the fixed-trajectory approach, errors that arose in previous sessions are never corrected and remain as they are. Therefore, the error continues to accumulate across the entire map as more sessions are added. (b) shows the result of optimizing all sessions with an anchor set per session. In this approach, the covariance of the anchor node is set much larger than the covariance of the intra-session factors. During optimization, the optimizer tends to reduce the inter-session loop residual by moving the low-cost anchor rather than deforming the costly intra-session structure. As a result, each session's trajectory effectively moves as a single rigid body, while the per-pose drift within it remains unresolved. Moving as a single rigid body fails to satisfy inter-session loop constraints that require different corrections at different points along the trajectory. As a result, residual alignment errors remain. In contrast, (c) places a large-covariance anchor at every submap and allows free trajectory deformation between submaps. As a result, the entire trajectory can move as a chain of multiple small rigid bodies. The inter-session loop constraints that demand different corrections at different locations can then be satisfied locally at the submap level. So, (c) achieves the consistent alignment result shown in the figure.

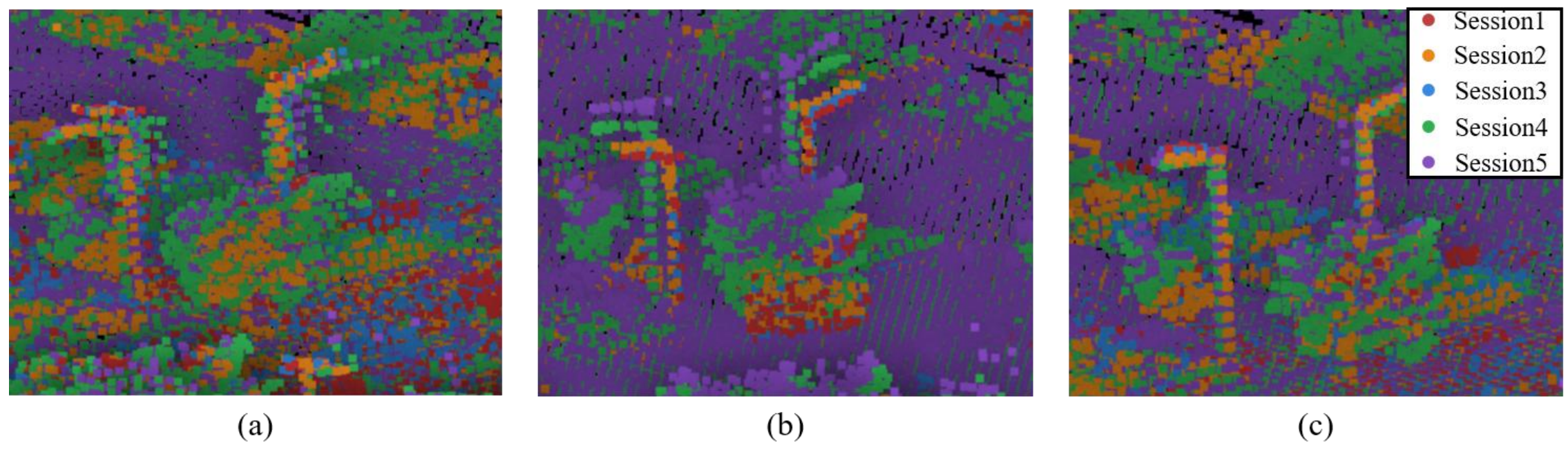

Fig. 5. Comparison of map alignment results after optimization of five sessions, obtained by (a) optimizing only the current session while keeping the trajectories of previous sessions fixed, (b) using only session-wise anchors for optimization and (c) using the proposed submap anchors for optimization.

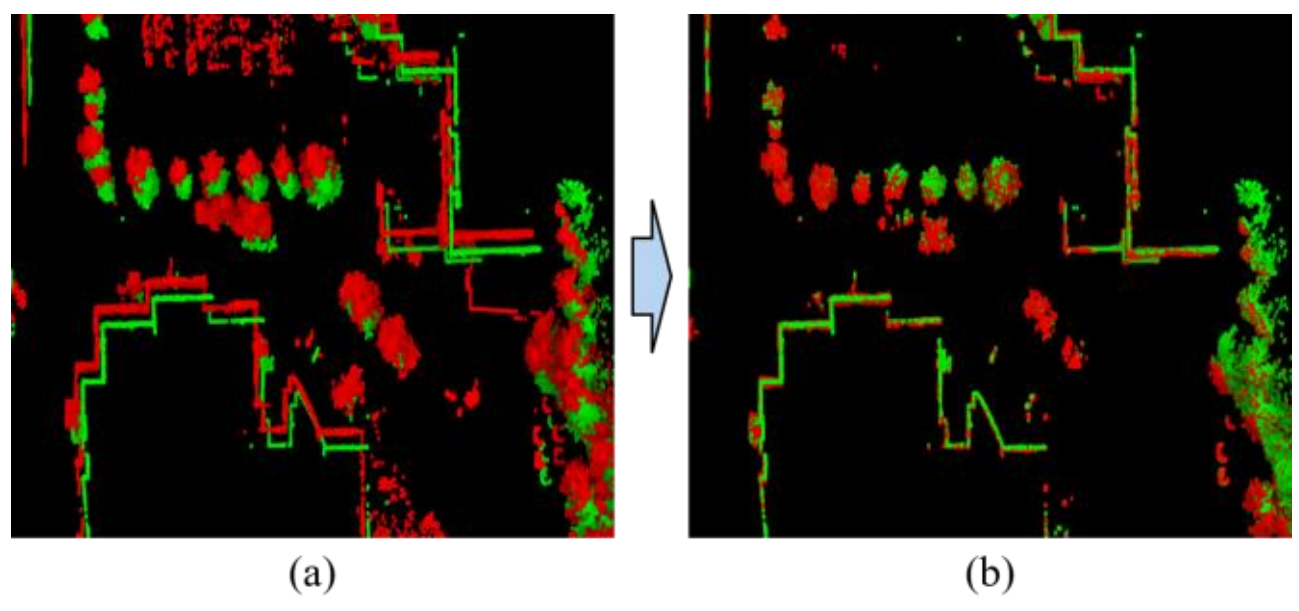

Fig. 6. Comparison of evidence reprojection results. Green points represent the previous-session point cloud optimized after session addition. Red points represent the evidence point cloud generated from the previous session. (a) is before reprojection and (b) is after reprojection.

TABLE II. COMPARISON OF RIGID ALIGNMENT RESIDUALS BY ANCHORING SCHEME ACROSS ACCUMULATED SESSION STEPS

| Session Step | Session Anchor | Submap Anchor |
|---|---|---|
| 01+02 | 890mm | 12mm |
| +03 | 872mm | 12mm |
| +04 | 1420mm | 19mm |
| +05 | 1780mm | 21mm |

Fig. 6 shows the result of reprojecting the evidence point cloud from a previous session onto the trajectory optimized after a new session is added. Before reprojection, the evidence point cloud from the previous session is misaligned with the optimized point cloud After reprojection, the evidence point cloud aligns with the optimized trajectory. Table II shows the pose residual before and after optimization, computed based on Umeyama rigid alignment [40]. The residual was measured using session data generated from the KAIST sequence of MulRan and HeLiPR which are large-scale driving datasets exceeding 6 km per session. The rigid registration residual of the session anchor approach keeps increasing as sessions accumulate. Because correcting the entire session with a single global anchor cannot sufficiently capture the trajectory deformation optimized over such a large-scale driving segment. In contrast, the Submap Anchor approach places multiple anchors at the local submap level and corrects the trajectory deformation of each segment individually. The submap anchor approach reduces the residual by more than 98% compared to the session anchor approach, regardless of the session stage.

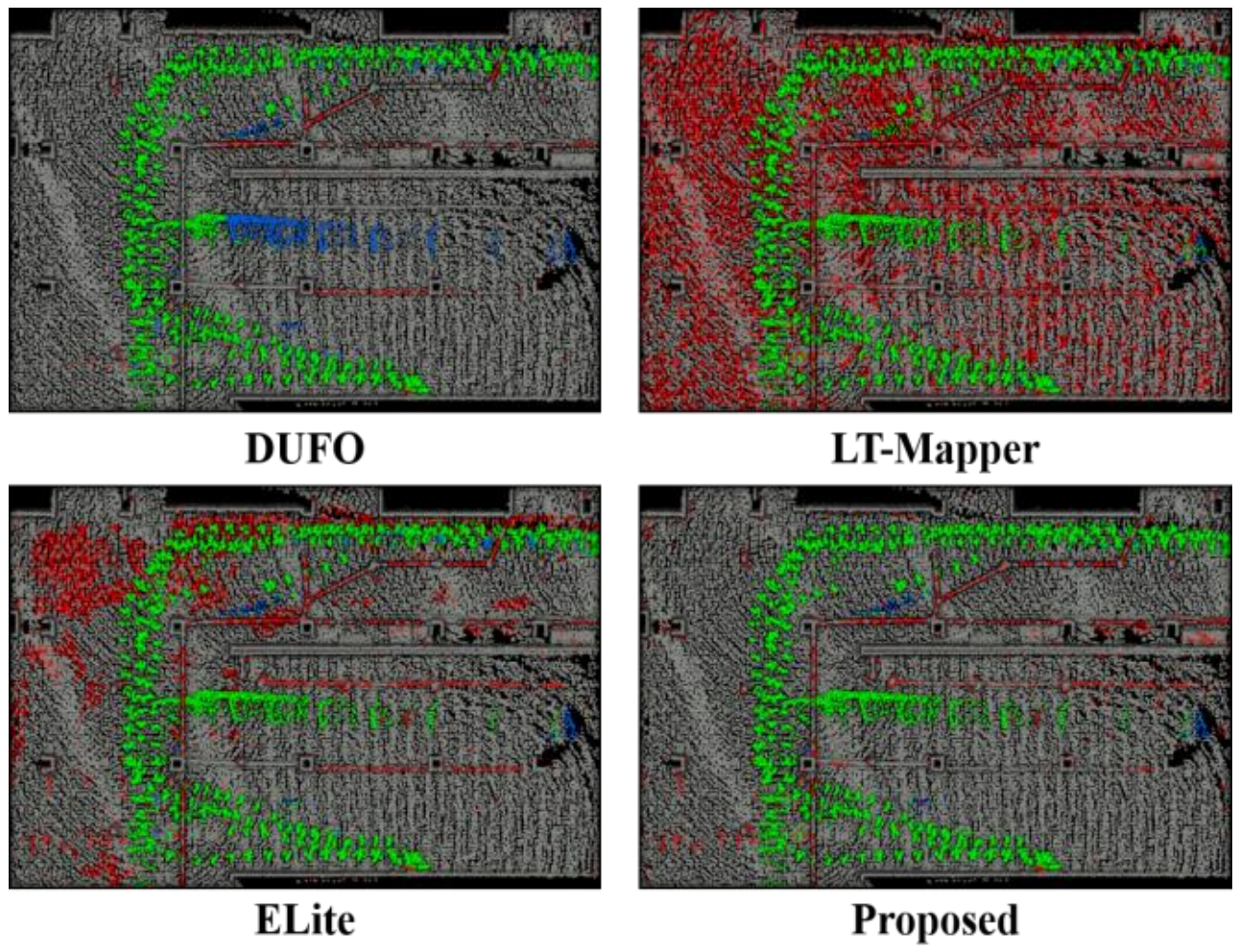

Fig. 7. Comparison of dynamic object removal result in the construction site dataset: True Positive (Green), False Negative (Blue), False Positive (Red).

TABLE III. DYNAMIC OBJECT REMOVAL RESULT ON CONSTRUCTION SITE DATASET

| Method | Session 1 | | | Session 2 | | | Session3 | | |
|---|---|---|---|---|---|---|---|---|---|
| | PR | RR | F1 | PR | RR | F1 | PR | RR | F1 |
| DUFOMap [12] | 87.14 | 92.37 | 89.68 | 91.16 | 96.36 | 93.69 | 95.07 | 89.47 | 92.19 |
| LT-Mapper [31] | 57.35 | 96.65 | 71.99 | 46.07 | 99.06 | 62.89 | 62.85 | 98.99 | 76.89 |
| ELite [19] | 91.50 | 96.76 | 94.06 | 84.26 | 97.80 | 90.53 | 86.96 | 97.04 | 91.73 |
| Proposed | 96.05 | 97.70 | **96.87** | 96.10 | 99.47 | **97.75** | 96.13 | 99.53 | **97.80** |

### B. *Dynamic Object Removal*

Fig. 7 compares DOR performance on the construction site dataset. We extracted ground truth through manual labeling and visualized TP, FN, and FP by color. Table III quantitatively compares each method using Precision (PR), Recall (RR), and F1-score. All methods show high RR indicating that dynamic object removal was performed well. However, the relatively low PR indicates that static structures were frequently misclassified as dynamic objects and removed. In contrast, the proposed method achieved the highest PR and the highest F1-score across all sessions. Existing methods are vulnerable to single-viewpoint information or a small number of observations. On the other hand, the proposed method conservatively adjusts the determination using hit and free evidence accumulated from multiple observations together with saturated support reliability. Therefore, it effectively reduces false positives on static structures compared to existing methods.

### C. *Map Update*

Fig. 8 shows the process by which the lifelong map and the final static map are generated as sessions progress. A gap of about four years occurred between Session 3 and Session 4.

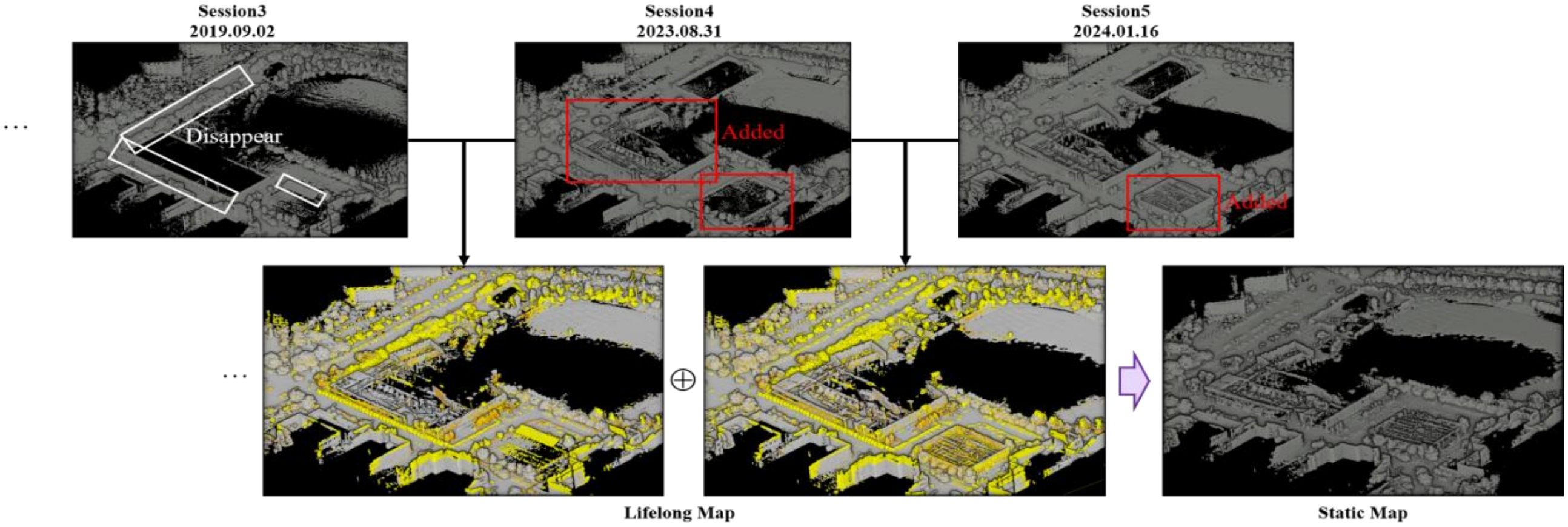

Fig. 8. Results of lifelong map and final static map generation over session progression. Structures that disappeared or newly appeared due to long time gaps between sessions are cumulatively marked as environmental change points (yellow) in the lifelong map. The final static map is generated by removing these change points, retaining only static structures.

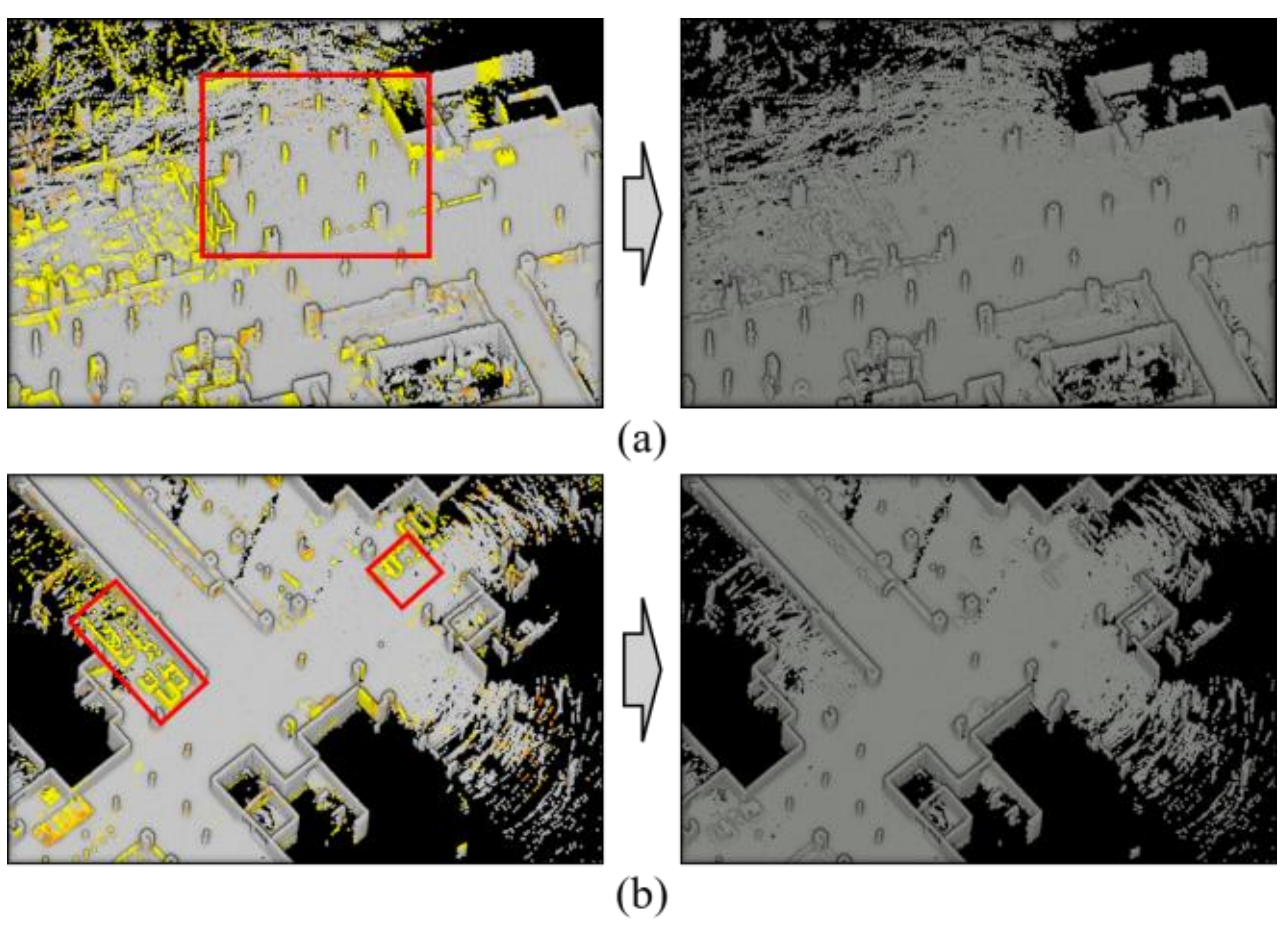

Fig. 9. Results of lifelong map (left) and static map (right) generation on the construction site dataset. (a) Temporary structures such as shoring and (b) temporarily stored construction materials. These points are detected as environmental change points and shown to be removed in the static map.

Structures that disappeared or newly appeared during the four-year gap are marked in yellow as points where environmental change occurred. The environmental change points accumulate as sessions progress. The static map, which retains only static structures, removes the accumulated change points. In addition, as shown in Fig. 9, temporary structures and construction materials are also reflected as environmental change points. The static map confirms the removal of the temporary structures and construction materials as well.

Table IV compares the processing time on the construction site dataset. The proposed method achieves the fastest total processing time. ELite performs forward and backward bidirectional GICP registration in the alignment stage. In the DOR stage, ELite splits each ray into individual points and performs a k-NN search to update $\varepsilon_l$ . Splitting each ray into individual points and performing a k-NN search makes ELite's processing time very slow. LT-mapper merges maps quickly. However, it takes a long time in the map update stage where HD objects are removed based on multi-resolution range images and inter-session changes are detected through KD-tree based nearest-neighbor search. In contrast, the proposed method reuses data from previous sessions and generates evidence only for the new session through submap-anchor-based evidence reprojection. In addition, it updates $\varepsilon_l$ through voxel-hash based lookup instead of point-wise k-NN search. These improvements together enable faster processing than the other methods.

TABLE IV. Processing Time of Each Method using Construction Site Dataset

| Method | Processing Time (sec) | | |
|---|---|---|---|
| | Merge | Update | Total |
| LT-Mapper [31] | 8.23 | 172.84 | 181.07 |
| ELite [19] | 409.32 | 1361.16 | 1770.48 |
| Proposed | 13.52 | 81.11 | 95.23 |

## V. Conclusion

We propose SEAM, a lifelong LiDAR mapping framework that jointly optimizes multi-session trajectories and reuses DOR and CD evidence at the submap level. SEAM reuses the evidence through reprojection after re-optimization, instead of recomputing the evidence from scratch. The DOP-based reliability metric, normalized by the LiDAR FoV, suppresses the generation of incorrect inter-session loop edges and thereby prevents errors in multi-session alignment. The direction aware voxel-wise evidence model reduces false detections in DOR and CD and improves accuracy. In experiments on a long-term multi-session dataset and a construction site dataset with severe environmental changes, the proposed framework maintains higher accuracy than existing methods that assume a fixed trajectory while substantially reducing processing time. The experimental results demonstrate that SEAM is a practical and scalable framework for lifelong mapping.

## References

[1] J. Zhang and S. Singh, "LOAM: Lidar odometry and mapping in real-time," in *Robotics: Science and Systems*, vol. 2, no. 9, 2014, pp. 1–9.

[2] T. Shan and B. Englot, "LeGO-LOAM: Lightweight and ground-optimized lidar odometry and mapping on variable terrain," in *2018 IEEE/RSJ Int. Conf. Intell. Robots Syst. (IROS)*, 2018, pp. 4758–4765.

[3] Shan, T., Englot, B., Meyers, D., Wang, W., Ratti, C., & Rus, D. (2020, October). Lio-sam: Tightly-coupled lidar inertial odometry via smoothing and mapping. In *2020 IEEE/RSJ international conference on intelligent robots and systems (IROS)* (pp. 5135-5142). IEEE.

[4] Xu, W., & Zhang, F. (2021). Fast-lio: A fast, robust lidar-inertial odometry package by tightly-coupled iterated kalman filter. *IEEE Robotics and Automation Letters*, *6*(2), 3317-3324.

[5] Guadagnino, T., Mersch, B., Gupta, S., Vizzo, I., Grisetti, G., & Stachniss, C. (2025, October). KISS-SLAM: A simple, robust, and accurate 3D LiDAR SLAM system with enhanced generalization capabilities. In *2025 IEEE/RSJ International Conference on Intelligent Robots and Systems (IROS)* (pp. 5363-5370). IEEE.

[6] Biber, P., & Duckett, T. (2005, June). Dynamic maps for long-term operation of mobile service robots. In *Robotics: science and systems* (pp. 17-24).

[7] Tipaldi, G. D., Meyer-Delius, D., & Burgard, W. (2013). Lifelong localization in changing environments. *The International Journal of Robotics Research*, *32*(14), 1662-1678.

[8] Vega-Torres, M. A., Braun, A., & Borrmann, A. (2024). SLAM2REF: Advancing long-term mapping with 3D LiDAR and reference map integration for precise 6-DoF trajectory estimation and map extension. *Construction Robotics*, *8*(2), 13.

[9] Wang, L., Zhong, X., Xu, Z., Chai, K., Zhao, A., Zhao, T., ... & Gao, F. (2026). Lemon-mapping: Loop-enhanced large-scale multi-session point cloud merging and optimization for globally consistent mapping. *IEEE Transactions on Automation Science and Engineering*.

[10] Lim, H., Kim, D., & Myung, H. (2025). Multi-Mapcher: Loop Closure Detection-Free Heterogeneous LiDAR Multi-Session SLAM Leveraging Outlier-Robust Registration for Autonomous Vehicles. *IEEE Transactions on Intelligent Vehicles*.

[11] Hornung, A., Wurm, K. M., Bennewitz, M., Stachniss, C., & Burgard, W. (2013). OctoMap: An efficient probabilistic 3D mapping framework based on octrees. *Autonomous robots*, *34*(3), 189-206.

[12] Duberg, D., Zhang, Q., Jia, M., & Jensfelt, P. (2024). DUFOMap: Efficient dynamic awareness mapping. *IEEE Robotics and Automation Letters*, *9*(6), 5038-5045.

[13] Lim, H., Hwang, S., & Myung, H. (2021). ERASOR: Egocentric ratio of pseudo occupancy-based dynamic object removal for static 3D point cloud map building. *IEEE Robotics and Automation Letters*, *6*(2), 2272-2279.

[14] Jang, S., Lee, A. J., Nahrendra, I. M. A., & Myung, H. (2026). Chamelion: Reliable Change Detection for Long-Term LiDAR Mapping in Transient Environments. *IEEE Robotics and Automation Letters*.

[15] Kim, T., Kang, G., Kim, T. I., Song, S., & Ko, H. K. (2026). MTD-Map: Single-Stage Long-Term LiDAR Map Maintenance Framework via Mixture Transition Distribution. *arXiv preprint arXiv:2606.29469*.

[16] Kümmerle, R., Grisetti, G., Strasdat, H., Konolige, K., & Burgard, W. (2011, May). g 2 o: A general framework for graph optimization. In *2011 IEEE international conference on robotics and automation* (pp. 3607-3613). IEEE.

[17] Kaess, M., Johannsson, H., Roberts, R., Ila, V., Leonard, J. J., & Dellaert, F. (2012). iSAM2: Incremental smoothing and mapping using the Bayes tree. *The International Journal of Robotics Research*, *31*(2), 216-235.

[18] Sünderhauf, N., & Protzel, P. (2012, October). Switchable constraints for robust pose graph SLAM. In *2012 IEEE/RSJ International Conference on Intelligent Robots and Systems* (pp. 1879-1884). IEEE.

[19] Gil, H., Lee, D., Kim, G., & Kim, A. (2025, May). Ephemerality meets LiDAR-based lifelong mapping. In *2025 IEEE International Conference on Robotics and Automation (ICRA)* (pp. 3312-3319). IEEE.

[20] Konolige, K., & Bowman, J. (2009, October). Towards lifelong visual maps. In *2009 IEEE/RSJ International Conference on Intelligent Robots and Systems* (pp. 1156-1163). IEEE.

[21] Kim, G., & Kim, A. (2018, October). Scan context: Egocentric spatial descriptor for place recognition within 3d point cloud map. In *2018 IEEE/RSJ International Conference on Intelligent Robots and Systems (IROS)* (pp. 4802-4809). IEEE.

[22] Kim, H., Choi, J., Sim, T., Kim, G., & Cho, Y. (2024). Narrowing your fov with solid: Spatially organized and lightweight global descriptor for fov-constrained lidar place recognition. *IEEE Robotics and Automation Letters*, *9*(11), 9645-9652.

[23] Xu, X., Lu, S., Wu, J., Lu, H., Zhu, Q., Liao, Y., ... & Wang, Y. (2023). Ring++: Roto-translation invariant gram for global localization on a sparse scan map. *IEEE Transactions on Robotics*, *39*(6), 4616-4635.

[24] Lim, H., Kim, D., Shin, G., Shi, J., Vizzo, I., Myung, H., ... & Carlone, L. (2025, May). Kiss-matcher: Fast and robust point cloud registration revisited. In *2025 IEEE International Conference on Robotics and Automation (ICRA)* (pp. 11104-11111). IEEE.

[25] McDonald, J., Kaess, M., Cadena, C., Neira, J., & Leonard, J. J. (2011). 6-DOF multi-session visual SLAM using anchor nodes.

[26] Kim, G., & Kim, A. (2020, October). Remove, then revert: Static point cloud map construction using multiresolution range images. In *2020 IEEE/RSJ International Conference on Intelligent Robots and Systems (IROS)* (pp. 10758-10765). IEEE.

[27] Chen, X., Milioto, A., Palazzolo, E., Giguere, P., Behley, J., & Stachniss, C. (2019, November). Suma++: Efficient lidar-based semantic slam. In *2019 IEEE/RSJ international conference on intelligent robots and systems (IROS)* (pp. 4530-4537). IEEE.

[28] Wang, N., Guo, R., Shi, C., Wang, Z., Zhang, H., Lu, H., ... & Chen, X. (2025). SegNet4D: Efficient instance-aware 4D semantic segmentation for LiDAR point cloud. *IEEE Transactions on Automation Science and Engineering*, *22*, 15339-15350.

[29] Andreasson, H., Magnusson, M., & Lilienthal, A. (2007, October). Has something changed here? autonomous difference detection for security patrol robots. In *2007 IEEE/RSJ International Conference on Intelligent Robots and Systems* (pp. 3429-3435). IEEE.

[30] Fehr, M., Furrer, F., Dryanovski, I., Sturm, J., Gilitschenski, I., Siegwart, R., & Cadena, C. (2017, May). TSDF-based change detection for consistent long-term dense reconstruction and dynamic object discovery. In *2017 IEEE International Conference on Robotics and automation (ICRA)* (pp. 5237-5244). IEEE.

[31] Kim, G., & Kim, A. (2022, May). LT-mapper: A modular framework for LiDAR-based lifelong mapping. In *2022 International Conference on Robotics and Automation (ICRA)* (pp. 7995-8002). IEEE.

[32] Segal, A., Haehnel, D., & Thrun, S. (2009, June). Generalized-icp. In *Robotics: science and systems* (Vol. 2, No. 4, p. 435).

[33] K. Kim, "Scan Matching Confidence Evaluation for Robust LiDAR Odometry and Pose Graph Optimization," *Journal of Institute of Control, Robotics and Systems*, vol. 31, no. 11, pp. 1299–1306, 2025.

[34] Langley, R. B. (1999). Dilution of precision. *GPS world*, *10*(5), 52-59.

[35] Bosse, M., & Zlot, R. (2013, May). Place recognition using keypoint voting in large 3D lidar datasets. In *2013 IEEE international conference on robotics and automation* (pp. 2677-2684). IEEE.

[36] Massatt, P., & Rudnick, K. (1990). Geometric formulas for dilution of precision calculations. *Navigation*, *37*(4), 379-391.

[37] Demantké, J., Mallet, C., David, N., & Vallet, B. (2011, August). Dimensionality based scale selection in 3D lidar point clouds. In *Laserscanning*.

[38] Kim, G., Park, Y. S., Cho, Y., Jeong, J., & Kim, A. (2020, May). Mulran: Multimodal range dataset for urban place recognition. In *2020 IEEE international conference on robotics and automation (ICRA)* (pp. 6246-6253). IEEE.

[39] Jung, M., Yang, W., Lee, D., Gil, H., Kim, G., & Kim, A. (2024). HeLiPR: Heterogeneous LiDAR dataset for inter-LiDAR place recognition under spatiotemporal variations. *The International Journal of Robotics Research*, *43*(12), 1867-1883.

[40] Umeyama, S. (1991). Least-squares estimation of transformation parameters between two point patterns. *IEEE Transactions on pattern analysis and machine intelligence*, *13*(4), 376-380.